\documentclass[letterpaper, 10 pt, conference]{ieeeconf}  

\IEEEoverridecommandlockouts                              

\usepackage{amsmath,amsfonts}
\usepackage{algorithmic}
\usepackage{algorithm}
\usepackage{array}
\usepackage[caption=false,font=normalsize,labelfont=sf,textfont=sf]{subfig}
\usepackage{textcomp}
\usepackage{stfloats}
\usepackage{url}
\usepackage{hyperref}
\usepackage{verbatim}
\usepackage{graphicx}
\usepackage{cite}

\title{\LARGE \bf
Car-STAGE: Automated framework for large-scale high-dimensional simulated time-series data generation based on user-defined criteria
}

\author{Asma A. Almutairi$^{1}$, David J. LeBlanc$^{1}$ and Arpan Kusari$^{1*}$
\thanks{*Corresponding author}
\thanks{$^{1}$A.A. Almutairi, D.J. LeBlanc and A. Kusari are with University of Michigan Transportation Research Institute, 
        University of Michigan, 2901 Baxter Road, Ann Arbor, MI-48103.
        {\tt\small \{asmaalm, leblanc, kusari\}@umich.edu}}%
}

\begin{document}

\maketitle
\thispagestyle{empty}
\pagestyle{empty}

\begin{figure*}[!t]
\centering

\subfloat[]{\includegraphics[width=0.5\linewidth]{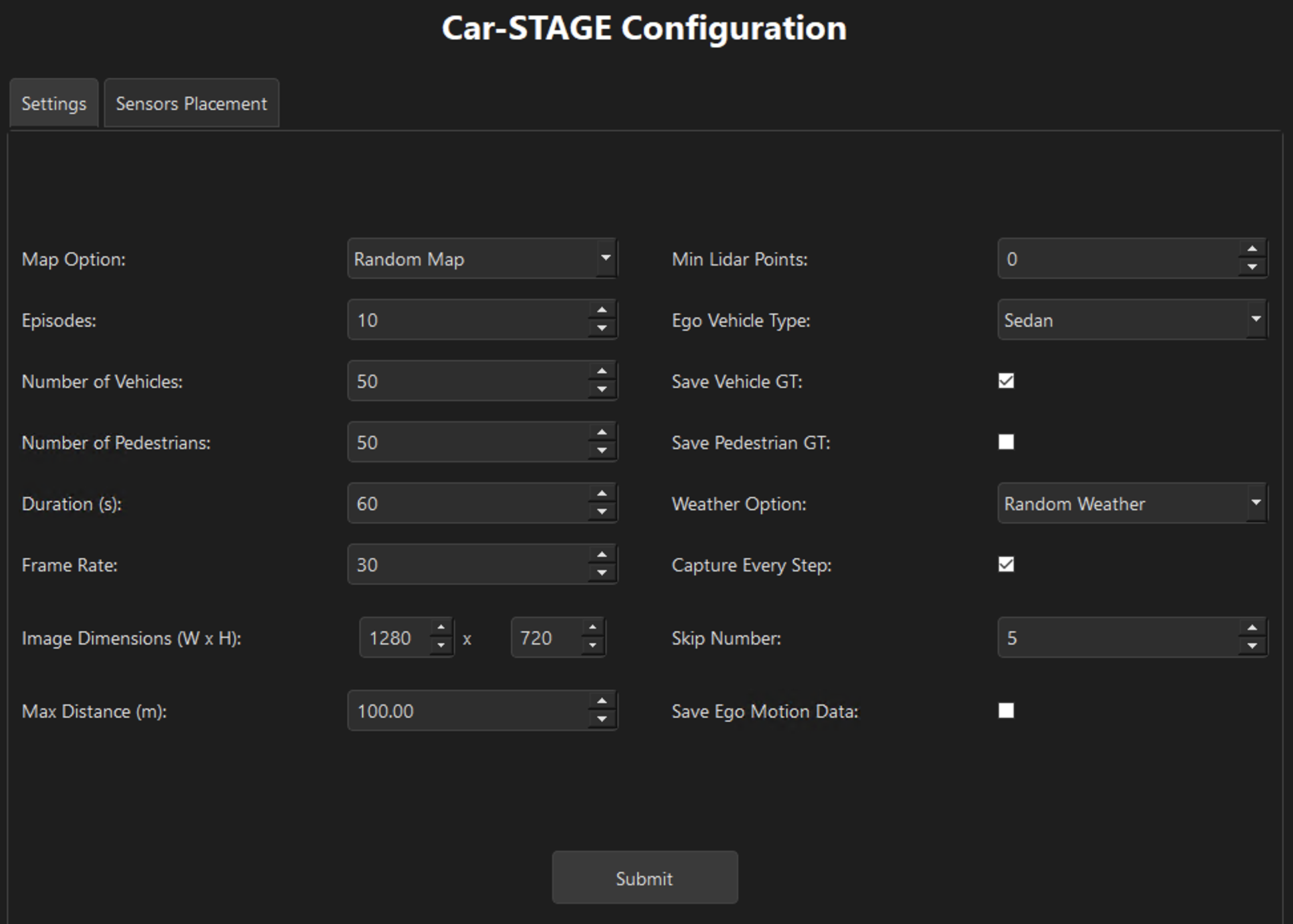}%
\label{fig_first_case}}
\hspace{0.02\linewidth}
\subfloat[]{\includegraphics[width=0.47\linewidth, height=2.5in]{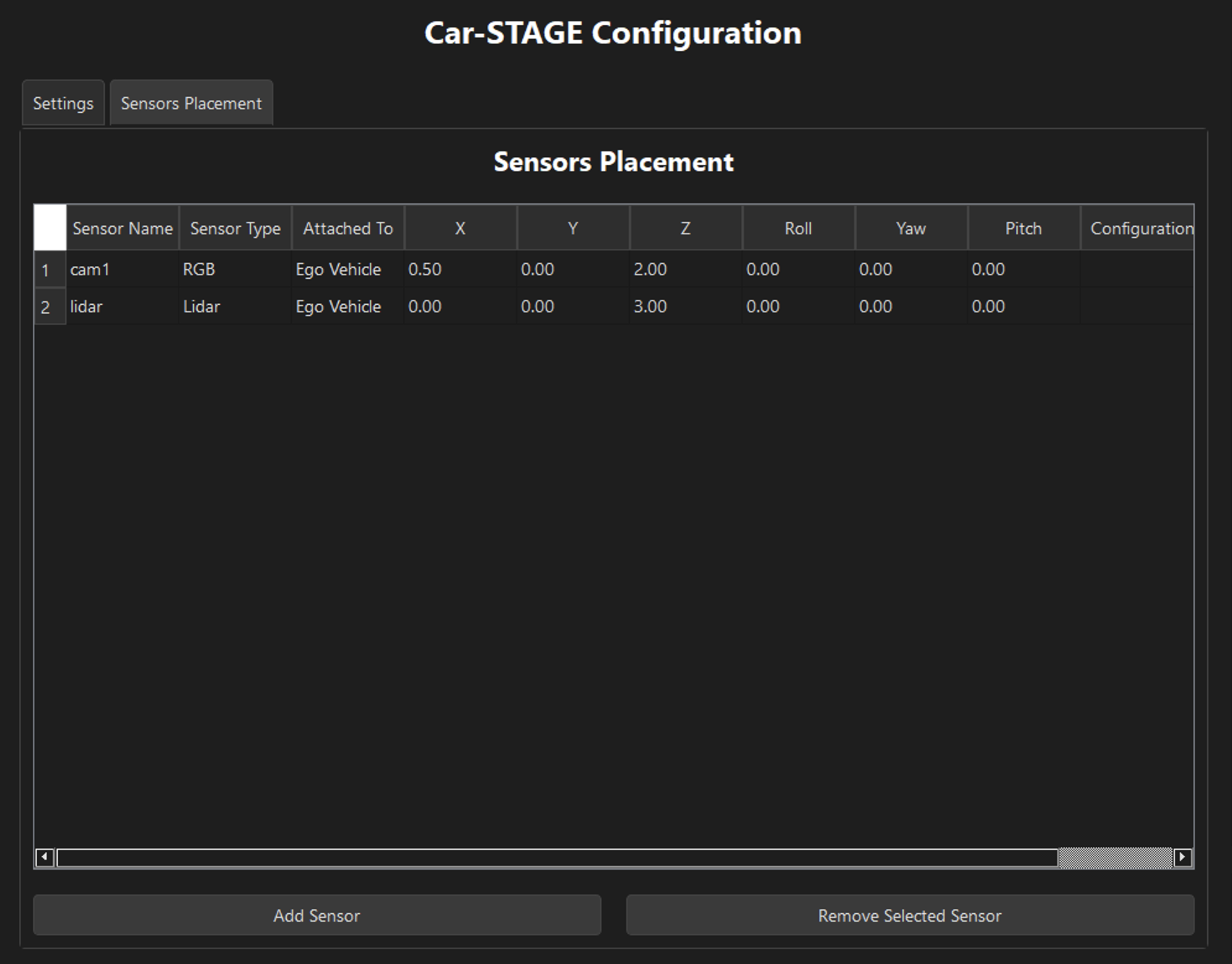}%
\label{fig_first_case}}
\caption{Screenshot of Car-STAGE GUI with the key parameters on the left and senors placement feature on the right}
\label{fig:gui}
\end{figure*}

\begin{abstract}

Generating large-scale sensing datasets through photo-realistic simulation is an important aspect of many robotics applications such as autonomous driving. In this paper, we consider the problem of synchronous data collection from the open-source CARLA simulator using multiple sensors attached to vehicle based on user-defined criteria. We propose a novel, one-step framework that we refer to as Car-STAGE, based on CARLA simulator, to generate data using a graphical user interface (GUI) defining configuration parameters to data collection without any user intervention. This framework can utilize the user-defined configuration parameters such as choice of maps, number and configurations of sensors, environmental and lighting conditions etc. to run the simulation in the background, collecting high-dimensional sensor data from diverse sensors such as RGB Camera, LiDAR, Radar, Depth Camera, IMU Sensor, GNSS Sensor, Semantic Segmentation Camera, Instance Segmentation Camera, and Optical Flow Camera  along with the ground-truths of the individual actors and storing the sensor data as well as ground-truth labels in a local or cloud-based database. The framework uses multiple threads where a main thread runs the server, a worker thread deals with queue and frame number and the rest of the threads processes the sensor data. The other way we derive speed up over the native implementation is by memory mapping the raw binary data into the disk and then converting the data into known formats at the end of data collection. We show that using these techniques, we gain a significant speed up over frames, under an increasing set of sensors and over the number of spawned objects. 

\end{abstract}


\section{INTRODUCTION}
\label{sec:intro}
The fields of Advanced Driver-Assistance Systems (ADAS) and Autonomous Vehicles (AV) rely heavily on extensive data-centric techniques to enable safe, efficient, and comfortable navigation through the dynamic environment. This data-driven approach is crucial for ADAS and AV systems to perceive their surroundings, anticipate potential hazards, and make informed decisions to ensure the safety and comfort of passengers and other road users. 

Recently, CARLA \cite{dosovitskiy2017carla} has come up as an open-source simulator for AVs which allows users to develop diverse scenarios to collect data and test various aspects of autonomous driving, such as urban traffic, highways and different environmental conditions. The users have control over maps and environmental conditions along with the different kinds and positions of sensors equipped in the subject vehicle (SV). In terms of operations, the native python API created by the CARLA team runs the server and client code in the same main thread process getting the server to generate a frame and then wait till client processes and saves the outputs. However, this makes the process inefficient and particularly slow. Secondly, CARLA can operate in asynchronous or synchronous manner - in the asynchronous manner, the server operates independently of the client while the synchronous mode leads to synchronizing between sensors using queue and frame number. However, the asynchronous mode can lead to data loss or data overrun, since there is no knowledge if sensors are working with data from same point in the simulation. On the other hand, the synchronous mode turns out to be really slow due to locking of the queue during I/O operations. 
Thirdly, the entire setup needs to be created by the users including synchronization of sensors and concurrent data collection strategies which presents an obstacle for even seasoned programmers.

Beset by these practical challenges, we present a novel, one-step framework, Car-STAGE (Fig.\ref{fig:gui}), based on the CARLA simulator that allows users to utilize a graphical user interface (GUI) to generate a collection of high-dimensional sensor data without any user intervention. This framework leverages the user-defined configuration parameters to autonomously run the simulation in the background, capturing sensor data from diverse modalities such as RGB Camera, LiDAR, Radar, Depth Camera, IMU Sensor, GNSS sensor, Semantic Segmentation Camera, Instance Segmentation Camera, and Optical Flow Camera, along with the ground-truth information of the individual actors within the simulated environment. We use a synchronous mode where we separate the data processing into a number of threads - a main thread which runs the server, a worker thread which deals with queue and frames and other executor threads that deal with sensors data. We also derive further speed up by saving raw data using memory mapping to disk. The collected sensor data and ground-truth labels are then stored in a local or cloud-based database, providing a comprehensive dataset for the development and evaluation of ADAS and AV algorithms. 

In Section \ref{sec:related}, we talk about related research utilizing CARLA simulation, while in Section \ref{sec:framework}, we provide the concepts of our proposed framework, Section \ref{sec:results} provides the comparisons in timings between our proposed framework and the native API on the basis of the number of frames, different number of sensors etc. Finally, Section \ref{sec:conclusions} provides some discussions and final conclusions. 




\section{RELATED WORK}
\label{sec:related}

There has been a few research works in the recent years which have utilized CARLA and are closest to our proposed framework. CarFree \cite{jang2021carfree} presents an open-source automatic ground-truth generation tool based on the CARLA simulator. CarFree consists of two main components: (i) a data extraction client that automatically gathers relevant information from the Carla simulator’s server, and (ii) a post-processing software that generates accurate 2D bounding boxes for vehicles and pedestrians in the collected driving images. Another work in the same vein is CARLA-GEAR \cite{nesti2022carla} which creates a tool for automatically creating photorealistic synthetic datasets for evaluation of adversarial robustness of neural models against physical adversarial patches. It also uses CARLA to create random patches in order to generate adversarial examples. \cite{rosende2023urban} provides aerial traffic generation dataset using images and labels of vehicles, pedestrians, and pedestrian crossings which the researchers provide for intelligent transportation management. None of the previously mentioned research focuses on enhancing the CARLA pipeline itself, which highlights the importance of the proposed work.

We would also like to note that there are other open-source simulators which have been used for AV simulation such as AirSim \cite{shah2018airsim} which is also built on Unreal Engine, similar to CARLA. However, CARLA has the most complete sensor set simulation, and better fidelity for defining complicated scenarios, which makes it the best choice for AV simulation.

\section{Proposed framework}
\label{sec:framework}

We name our proposed framework Car-STAGE, which stands for \textbf{Car}la-based \textbf{S}imulated \textbf{T}ime-series \textbf{A}utomated \textbf{G}eneration \textbf{E}ngine. The framework has been constructed upon the CARLA simulator, incorporates a GUI alongside a comprehensive suite of 12 modules. The GUI serves to facilitate user-driven configuration of simulation parameters, encompassing aspects such as duration, weather conditions, sensor specifications, synchronization modes for sensors, and labeling methodologies, while the modules are designed to facilitate environmental manipulation, object instantiation, and the extraction of metadata from the generated data, inclusive of annotations, thus guaranteeing the production of high-fidelity and precise data. The resultant data may be stored either in local storage solutions or on remote servers. The Car-STAGE framework is compatible with CARLA versions 0.9.13 and subsequent releases. 

\subsection{Project Setup}
\label{subsec:setup}
\begin{figure}[!t]
\centering
\includegraphics[width=1\linewidth]{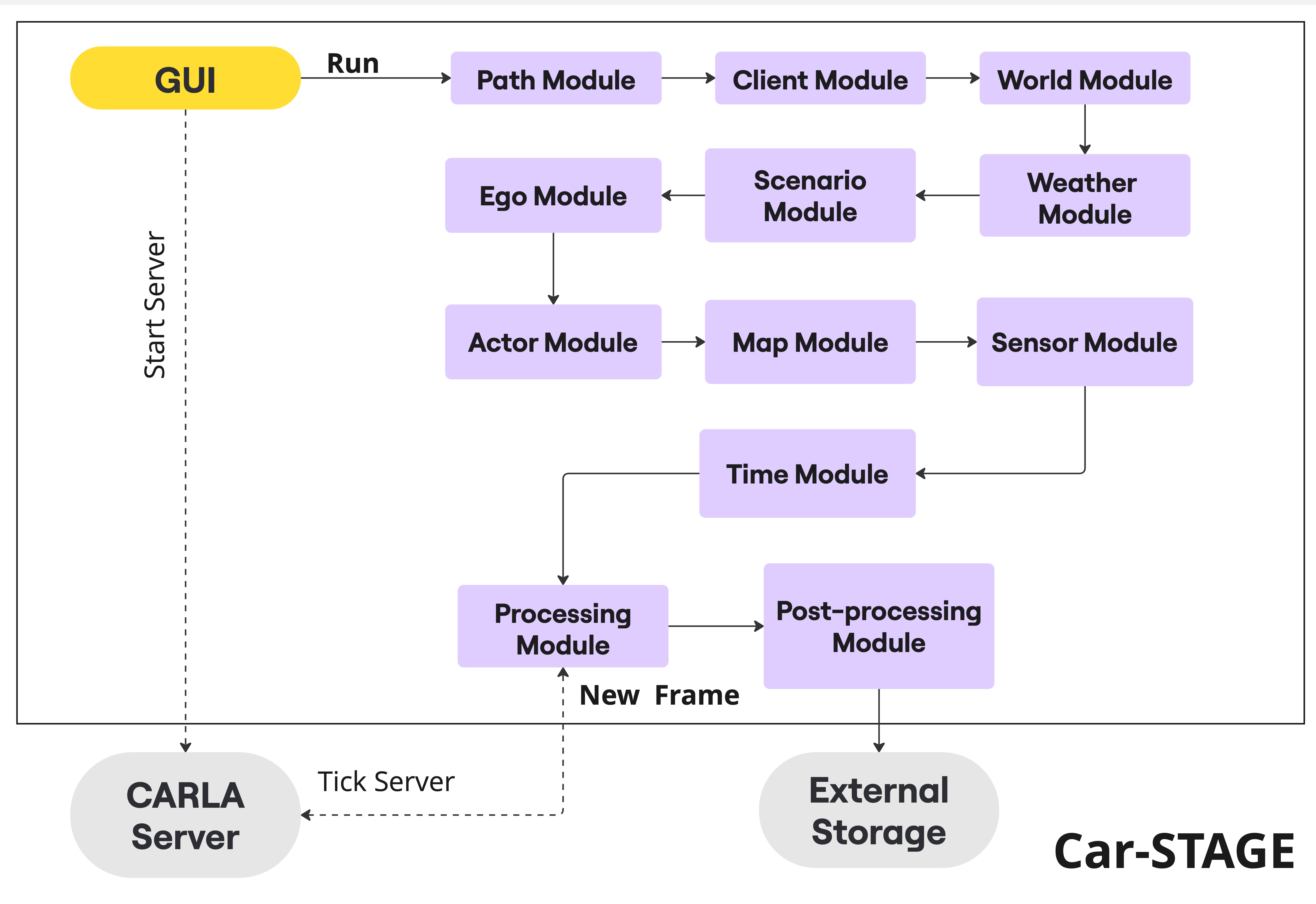}
\caption{Schematic representation of the various Car-STAGE modules}
\label{fig:modules}
\end{figure}

Car-STAGE consists of using the GUI to provide user-defined settings, and execute the program. It will then launch the CARLA server and conduct simulation episodes ("runs") based on the user's prior input, subsequently saving the output. The project is structured into two primary components: GUI and modules. The GUI, as shown in Fig. \ref{fig:gui}, allows the user to change Car-STAGE settings and will be elaborated upon in the subsequent section. CARLA consists of 12 modules, which we modify significantly to streamline operation and data generation without requiring users intervention.  Fig. \ref{fig:modules} illustrates the modules of Car-STAGE, which include the following: 1- Path module, which specifies the CARLA directory and the Car-STAGE output directory; 2- Client module, a wrapper class for CARLA client, designed for interfacing with the CARLA server. 3- World module, a wrapper class for the CARLA world to encapsulate the present state of the simulation environment; 4- Weather module, to alter the simulation weather using a specified selection of CARLA weather settings; 5- Scenario module,  delineates the behavior of simulation actors, defaulting to an autopilot mode in which each actor adheres to safe driving regulations as prescribed by the CARLA traffic manager object. 6- Ego module, designed to define the sort of ego vehicle and facilitate its spawning, with the resultant object subsequently utilized in conjunction with the sensor module to affix a collection of sensors on the ego vehicle; 7- Actor module for generating pedestrians and additional vehicles;  8- Map module for loading the CARLA map for simulation, configurable to either a random or particular map. 9- Sensor module for configuring sensors, establishing their position and orientation, mounting them to the ego vehicle, and initiating data streams from each sensor, presented as a feature on the right side of Fig. \ref{fig:gui} and elaborated in subsection \ref{subsec:sensor}; 10- Time module for configuring simulation and physical time, described in subsection \ref{subsec:sync};  11- Processing module analyzes data from the World and Sensor modules, detailed in subsection \ref{subsec:data_processing}. and 12- Post-processing output module which manages the storage of data to the designated destination, elaborated in subsection \ref{subsec:data_postprocessing}. The modular structure was established to accelerate the project's advancement, enabling several modules to operate independently without undermining the framework's integrity.

\subsection{Simulation Settings GUI}
\label{subsec:gui}
The GUI helps the users to configure the following parameters: 
\begin{itemize}
    \item option for random map or selection from a predefined list. 
    \item number of episodes (runs);
    \item number of vehicles;
    \item number of pedestrians;
    \item option to save ground-truth data of each vehicle in frame;
    \item option to save pedestrian ground-truth data in frame;
    \item simulation duration in seconds per run;
    \item frame rate; 
    \item image dimensions; 
    \item maximum distance from the ego vehicle for data capture; 
    \item minimum number of lidar points per object to consider visible in frame; 
    \item ego vehicle type; 
    \item option for random weather or selection from a predefined list;
    \item flag for capturing every step; and
    \item option to save ego motion data
\end{itemize}

\begin{figure}
    \centering
    \includegraphics[width=1\linewidth]{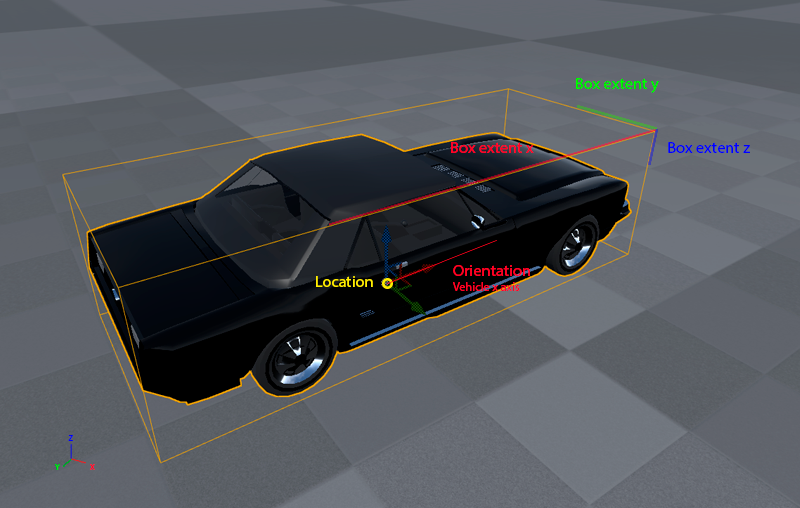}
    \caption{Bounding box measurements in CARLA. X,Y, and Z are center of bounding box. While extent x, extent y, extent z correspond to half the dimensions of the bounding box along the respective x, y, and z axes. (Image source: CARLA)}
    \label{fig:bounding_box}
\end{figure}

\subsection{Coordinate System}
\label{subsec:coord_system}
Car-STAGE uses the CARLA coordinate system (refer to Fig. \ref{fig:bounding_box}), which is congruent with the z-up left-handed configuration of Unreal Engine 4's coordinate framework. The positive X-axis points forward, the positive Y-axis points to the right, and the positive Z-axis points upwards. Positive steering induces a rightward rotation, while negative steering induces a leftward rotation. The rotation is characterized by pitch, yaw, and roll corresponding to Y-rotation, Z-rotation, and X-rotation respectively, in contrast to the configuration of the Unreal Engine (roll, pitch, yaw). The position and rotation attributes of entities such as ego vehicle, other vehicles, and pedestrians under the Car-STAGE framework are denoted by local coordinates.

\subsection{Sensors and Sensors Placement}
\label{subsec:sensor}
Car-STAGE supports multiple camera sensors: RGB cameras returning RGB images, depth cameras returning a coded depth image, semantic segmentation cameras classifying each object in an image by its tag, instance segmentation cameras (from CARLA 0.9.14) classifying each object by both tag and instance ID, and optical flow cameras capturing the motion perceived from the camera's point of view. It further includes a spinning LiDAR sensor that generates the point clouds; a radar sensor that creates points in close vicinity to the sensor while measuring distance, angle, and velocity; an IMU sensor supplying data from the accelerometer and gyroscope relative to the ego vehicle; and a GNSS sensor estimating the current GNSS position. All sensors operate in the UE coordinate system: x-forward, y-right, z-up, and output their coordinates in local coordinate. Attaching a sensor to a parent class in the CARLA simulation consists of two different functions, namely, attachment-to and attachment-type. The former gives the parent class which sensor attached to, could be the ego vehicle itself, other vehicles, or even the traffic lights, whereas the second one gives the relative position of the sensor w.r.t the parent class. Resulting in a three different attachment positions, namely: Rigid attachment-the movement is directly linked to its parent location and is, therefore, suitable for data retrieval from the simulation; SpringArm attachment-which allows for movement with small accelerations and decelerations and is mainly recommended in cases of video recording of the simulation due to its smooth movement during updates of the camera position; and SpringArmGhost attachment-similar to the previous setting but without collision testing, allowing the camera or sensor to pass through walls or other obstacles. Car-STAGE manages this process by creating all sensors that are to be mounted on the top of ego vehicle via a rigid body only. It offers a GUI for placing sensors (refer to Fig. \ref{fig:gui}), and this allows for specifying an x and y positions and also roll, pitch and yaw for each sensor. It also makes easy modification of other sensor settings such as the field of view (FOV). 

\subsection{Synchronized Multi-sensors Data Collection}
\label{subsec:sync}
 Car-STAGE incorporates four fundamental concepts derived from CARLA: (a) simulation time which defines the progression of time in the simulation, (b) physics sub-stepping which defines the precision of actors physics calculations, (c) client-server synchronization which defines the method by which the server advances the simulation in conjunction with the client, and (d) synchronization of sensors. These concepts are crucial for ensuring determinism and data integrity within Car-STAGE simulation. 
 
\subsubsection{Simulation Time}
\label{subsubsec:sim_time}
The Car-STAGE simulation environment distinguishes between simulation time and real time, as they differ fundamentally in nature and application. 
Simulation time is the internal clock of a virtual environment that determines the timing and sequence of events within the simulated world. 
In contrast, real time refers to the actual time elapsed in the physical world during the simulation process. One unit of simulation time does not need to correspond to one unit of real time - the power of hardware, and the quality of code determines how much time it takes to generate one frame. Another important temporal consideration is the length of time between two consecutive frames that are simulated. This can be set to either a variable or fixed timestep. By default, CARLA runs at a variable timestep mode that represents the real time spent by the server, computing every subsequent frame. It may differ between different frames within the same run and auto-adjusts itself depending on the computational load, hence having anomalies. Fixed time step mode in Car-STAGE ensures uniformity in rendering the frames and reliability in their calculation by maintaining a constant time difference between frames using $\text{Fixed Timestep} = \frac{1}{FPS}$ where FPS stands for frames per second. Therefore, for a user-specified frame rate of 30 FPS, Car-STAGE produces 30 frames every second of simulation time as opposed to real time. 



\subsubsection{Physics sub-stepping}
\label{subsubsec:substepping}
In general, simulations using physics engines have to calculate updates more frequently to keep the physics stable and accurate in virtual worlds. CARLA also supports this by physics sub-stepping, in which the engine can carry out multiple physics calculations in a single frame to keep it consistent, independent of frame rates. For lower frame rates, the number of physics updates per frame grows proportionally to preserve the quality of the simulation. That sub-stepping mechanism depends upon two critical parameters: Max Substep Delta Time and Max Substeps. Max Substep Delta Time determines the maximum amount of time in seconds per physics sub-step. In order to achieve a physics update rate of 120 Hz, this would be set to 1/120 seconds, or approximately 8.33 milliseconds. This will ensure that the physics timestep does not go over this value, which will maintain simulation stability even when simulation frame rates go over this threshold. Max Substeps is the maximum number of physics sub-steps allowed between two render frames. If the value is set to 4, for instance, the engine can make up to four discrete physics calculations per frame, if needed. The interplay of these parameters with the actual frame rate at a given time determines the physics calculation methodology. Sub-stepping is not required if the frame rate exceeds the physics engine's operational requirements. For example, if a simulation ran at 120 FPS and the physics engine required a minimum of 100 FPS to act optimally, then at this rate, physics updates can occur one-to-one or with slightly slower simulation, without the need for physics sub-stepping. However, if the frame rate of the same simulation were to decrease by half, the physics engine requirements would be outpaced by the simulation frame rate. In this case, physics sub-stepping would allow for higher calculations to provide equivalent physics computation as if the simulation were running at 120 FPS. To ensure the physics engine's functionality and accuracy with respect to frame rate, Car-STAGE enforces $\text{Fixed  Timestep} <= msdt * ms$, where msdt represents Max subset Delta Time and ms denotes Max Substeps. 

\subsubsection{Client-server synchronization}
\label{subsubsec:client_sync}
As mentioned in Section \ref{sec:intro}, CARLA can operate both in asynchronous and synchronous modes. However, the default setting is asynchronous where the server operates on its own and independently of the client. This can lead to complication, the server may process data at a rate which is faster than that of the client, which may cause the loss of data or data overrun. It becomes very difficult to ensure that every sensor is coordinated to the same moment within the simulation. In contrast, synchronous mode in CARLA is problematic since the operations all happen in the same thread which means that the server remain idle until it receives a "tick" signal from the client before it can proceed to process the subsequent frame. 

Car-STAGE implements the synchronous mode of operations by default where a single client spawns multiple threads, but only one thread assumes the responsibility of managing the data flow by dispatching tick messages to the server. This methodology ensures the precise collection of all data.

\subsubsection{Sensor synchronization}
\label{subsubsec:sensor_sync}
Synchronization of sensors is very critical to allow the acquisition and processing of sensor data to be perfectly aligned with the simulation timeline, hence emulating the real-time characteristics of the physical sensors in an AV. GPU-based sensors, such as cameras, would have several frame delays in most of the situations. It is necessary to apply synchronization, queuing and matching frame numbers in case some precision is needed; otherwise, the generation of sensor data can be done independently and each sensor would work at its own pace. In Car-STAGE, this is configured for both scenarios, with the default setting being synchronous for sensor output. Overall, Car-STAGE was developed to operate with a fixed time-step, incorporating valid physics sub-stepping, client-server synchrony, and synchronized sensors. This configuration was determined to be optimal for determinism, precision, and reproducible results.

\subsection{Data Processing}
\label{subsec:data_processing}
\subsubsection{Sensors Data}
At the beginning of each episode, Car-STAGE creates an object for each sensor of the respective class, whose methods include generating data and putting these data into a queue. The sensor module then sets up a list that includes all sensor queues. During execution, as the client sends a tick to the server to render the next frame in the simulation, the client receives the response of the server, which includes the new number of the new frame in the virtual world. Car-STAGE uses that new number to index in the sensor data queue compilation and checks that the relevant sensor data are all marked with the same frame number. If inconsistencies are detected, Car-STAGE stops further processing and moves on directly to the next frame. This would only happen if and only if sensor synchronization had been established for the sensors earlier through the GUI of Car-STAGE. The output of this step is a frame data object for each sensor. This object will be utilized to process and store data. We save the raw binary data of each camera sensors to disk using memory-mapped file, where the size of each memory-mapped file = 1 second (batch duration) * FPS * Image width * Image height * Channel number. This object will also be used for further processing to obtain annotation data. All I/O tasks in Car-STAGE are operated by a pool of multiple threads.

\subsubsection{Annotation}
Each actor in CARLA has  location  (x,y,z) and extent  (x,y,z), which together form a 3D bounding box that encapsulates the actor, as illustrated in Fig. \ref{fig:bounding_box}.  This bounding box is initially defined in the actor’s local coordinate frame and must be transformed into the world coordinate system before being further converted into the sensor coordinate system for accurate spatial representation.  By default, CARLA does not differentiate between actors positioned in front of the ego vehicle or sensors and those that are far away or occluded by large objects, such as buildings.  Also, for every generated frame, CARLA provides a list of all actors within the simulation world, regardless of their relative position or visibility to the sensors. To address this limitation, Car-STAGE introduces the Car-STAGE Visible Objects (STAGE-VO) algorithm. 


The algorithm was inspired by the methodology proposed in \cite{mukhlas2020carla}. The process begins by filtering out actors that are beyond a predefined distance from the sensor, then passes the rest of the actors to is\_visible function. This function takes as input the ego vehicle object, sensor object, depth image, extrinsic, max distance, and sensor parameters. The visibility determination is performed in two primary steps. First, the function computes the 2D bounding box of an object in the sensor frame by projecting the object's 3D bounding box into sensor frame. Second, the depth map is used to verify whether the projected bounding box is occluded. The depth image is  converted from a three-channel RGB encoding to a single-channel grayscale depth map using Eq. \ref{eq:encoding}.

\begin{align}
D =  1000 \times \frac{R + G \cdot 256 + B \cdot 256^2}{256^3 - 1}
\label{eq:encoding}
\end{align}

Using this depth image and the projected 2D bounding box, occlusion statistics are then computed for the set of 2D projected vertices in the sensor data. For each projected vertex (u,v), the function verifies whether the actual depth from the depth map at that pixel is smaller than the computed depth Z from the projection. If it is smaller than the vertex, then it is occluded; otherwise, it is classified as visible. The number of visible vertices is calculated to categorize occlusion levels into three classes: 0 (Fully visible):  At least six vertices are visible. 1 (Partially occluded):  Four or five vertices are visible. 2 (Mostly occluded):  Fewer than four vertices are visible. After that, for each bounding box, the object's orientation relative to the sensor is computed by determining two angles: the rotation angle (Eq. \ref{eq:yaw}) -

\begin{align}
\theta_y = \text{yaw}_{\text{o}} - \text{yaw}_{\text{s}} - 90^\circ
\label{eq:yaw}
\end{align}

where \( \text{yaw}_{\text{o}} \) represents the yaw angle of the object, \( \text{yaw}_{\text{s}} \) represents the yaw angle of the sensor and the subtraction of \( 90^\circ \) aligns the yaw measurement with the sensor's coordinate system \cite{astudillo2022mono} and the observation angle (Eq. \ref{eq:obs_angle}) -

\begin{align}
\alpha = \theta_y - \arctan \left(\frac{x}{z} \right)
\label{eq:obs_angle}
\end{align}
where \( x, z \) are the object's relative coordinates in the sensor frame.

Finally, the truncation ratio for each bounding box is computed by subtracting eight from the number of visible vertices obtained during the occlusion calculation. The resulting value is then divided by eight to obtain the truncation ratio.

\subsection{Data Post-processing}
\label{subsec:data_postprocessing}
After all runs/episodes have been executed and completed, the server stops running and all threads complete their jobs, the post-processing phase begins. We want to save the sensor objects into common formats such as png image files. Thus, we run a process pool and run through the memory-mapped files and convert it into common formats for the given sensor dat. 

\section{Results}
\label{sec:results}

\begin{figure}
    \centering
    \includegraphics[width=1.0\linewidth]{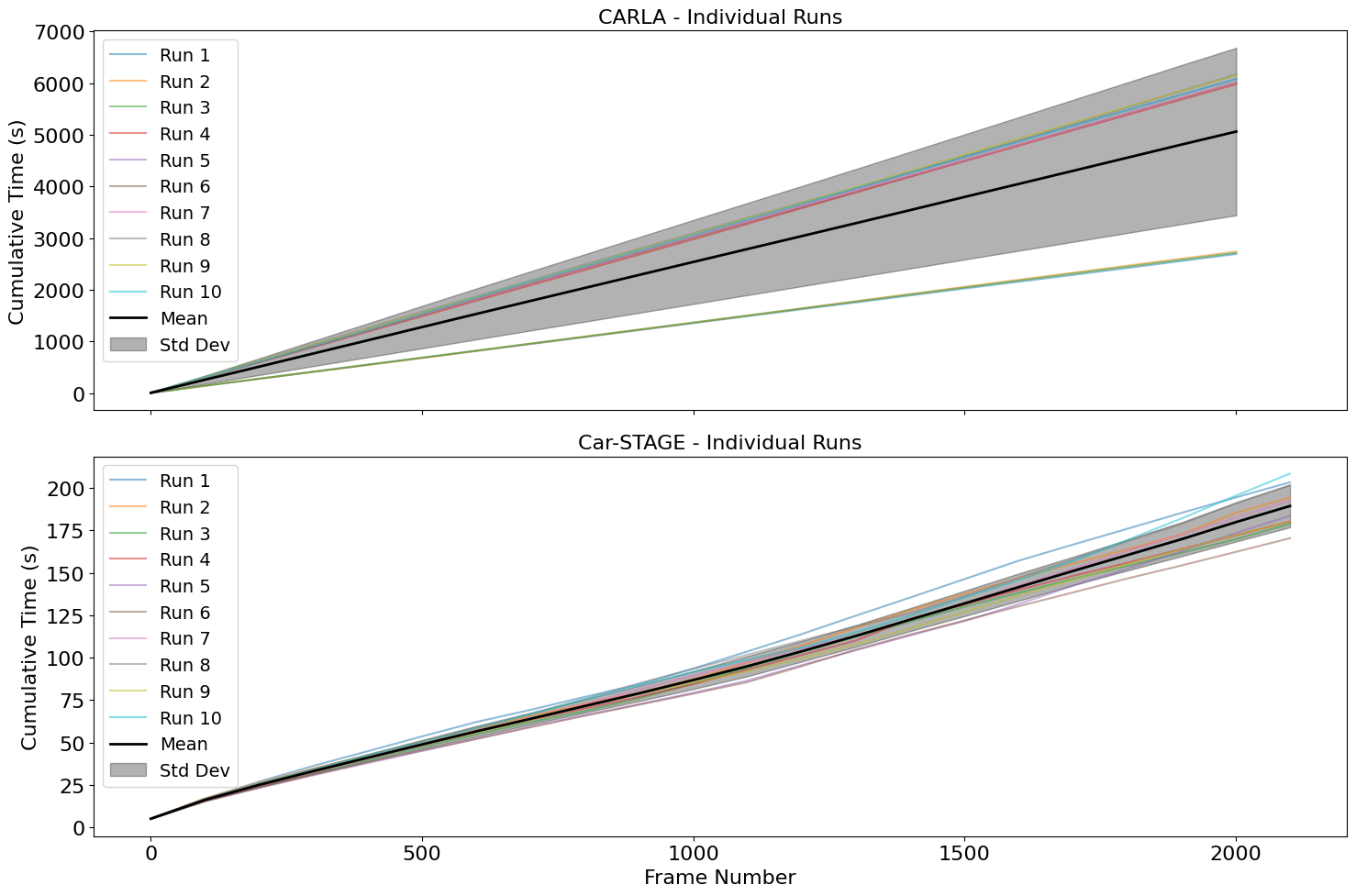}
    \caption{Comparison of total time (in sec) for CARLA (above) and Car-STAGE (below) over the number of frames}
    \label{fig:timing_num_frames}
\end{figure}

\begin{figure}
    \centering
    \includegraphics[width=1.0\linewidth]{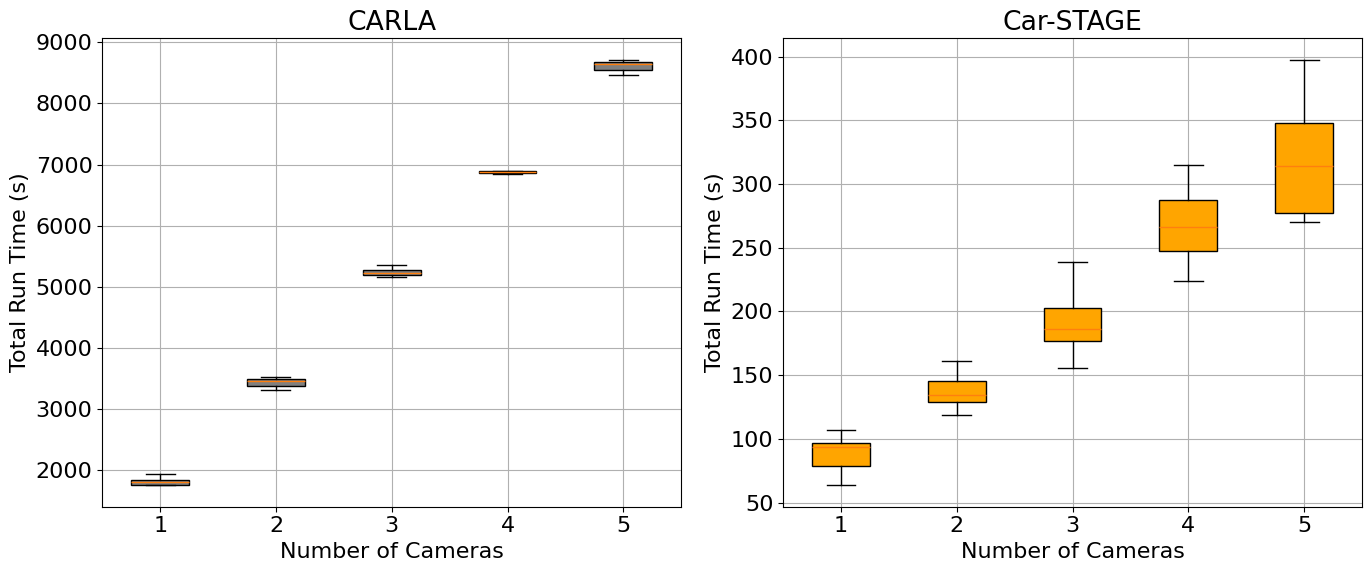}
    \caption{Comparison of total time (in sec) for CARLA and Car-STAGE as a function of the number of cameras}
    \label{fig:timing_num_cameras}
\end{figure}

\begin{figure}
    \centering
    \includegraphics[width=1.0\linewidth]{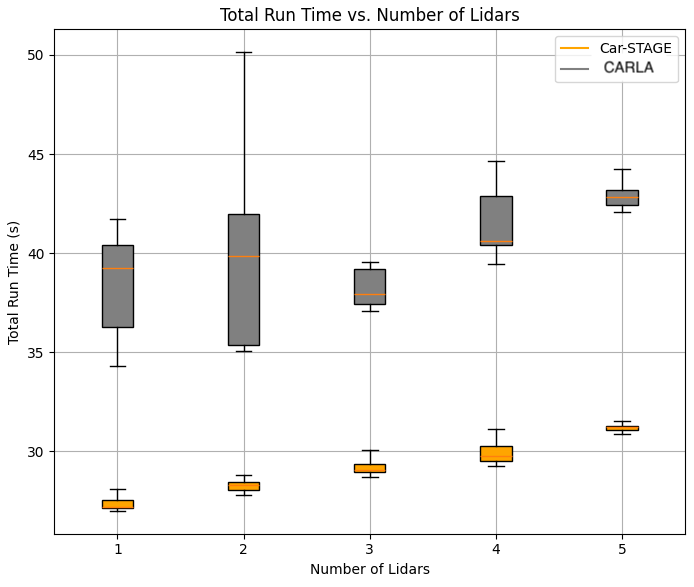}
    \caption{Box plots of comparison of total run-time (in sec) for baseline (native CARLA implementation) and Car-STAGE as a function of the number of LiDARs}
    \label{fig:timing_num_lidars}
\end{figure}

We aim to showcase the performance that our proposed framework provides over the native CARLA implementation in terms of number of frames, number of cameras and LiDARs spawned in each run. All the experiments are run in an Ubuntu workstation version 20.04.6 with 12th Gen Intel core i9-12900H processor with 20 threads and 3.77 GHz processing power and one graphics card with NVIDIA GeoForce RTX 3080, CUDA VERSION: 12.3, and Carla version 0.9.14.
We run each experiment over 10 identical runs and provide the mean and standard deviation of the results. 

Firstly, we run an experiment to provide a comparison between baseline CARLA and Car-STAGE in terms of number of frames output. For this experiment, we set the number of vehicles and pedestrians spawned to be 50 each, the frames per second to 30 and an assortment of sensors: 4 RGB cameras, 1 ray-cast LiDAR, 1 semantic LiDAR,  GNSS, IMU, depth camera, 1 semantic segmentation camera and 1 instance segmentation camera. Finally, we create bounding boxes for all vehicles and pedestrians for every frame. Fig. \ref{fig:timing_num_frames} shows the performance of CARLA and Car-STAGE over the frames. We find that while the increase in both CARLA and Car-STAGE are linear, the slope of Car-STAGE is far more gradually increasing as opposed to CARLA. The spread of timing over different runs for CARLA is also much larger than Car-STAGE which points towards allocation of processing resources in the single thread operation. 

Next, we provide the comparison between baseline CARLA and Car-STAGE with the number of frames fixed to 2000 and steadily increasing the number of cameras attached to the ego-vehicle. The image frame resolution is 1280x720 and the image format is Portable Network Graphic (PNG) format for both CARLA and Car-STAGE. Fig. \ref{fig:timing_num_cameras} shows the box plot comparison with the addition of a single camera to the processing upto 5 cameras. While increasing a camera to Car-STAGE modestly increases the amount of processing (about 50 sec for each camera), for CARLA, it leads to a 1500-2000 sec increase which is almost equivalent to  the I/O operations needed for 2000 frames. 

As a third case-study, we compare between baseline CARLA and Car-STAGE by increasing the number of LiDARs attached to the ego-vehicle and setting the number of frames to 2000. Fig. \ref{fig:timing_num_lidars} shows the box plot comparison between Python API of CARLA and Car-STAGE over the number of LiDARs. We find that on an average, Car-STAGE provides about 30\% speed up over the native implementation. We also find that there is a much larger spread in the native CARLA implementation rather than in Car-STAGE which shows the inconsistency in having a single thread processing the entire pipeline. Finally, we find that growth of time with respect to the increase in LiDAR is almost similar in both CARLA and Car-STAGE. 

\section{Conclusions}
\label{sec:conclusions}

We propose a scalable and robust data generation tool in this paper based on CARLA simulator. It solves the problem of large scale data collection that CARLA is designed for while providing a one-step GUI based tool for flexibility of users. The GUI provides various configuration options such as choice of maps, number of episodes, vehicles, pedestrians, weather conditions etc. Internally, Car-STAGE modifies the software modules of CARLA to streamline operations by focusing on a synchronous mode of data collection. Instead of a single thread operation that is used in CARLA, Car-STAGE uses a multi-threaded approach in which a main thread runs the server, a worker thread which deals with the queue and other executor threads which process sensor data. We show in results that using this model along with some easy and intuitive choices in data processing leads to substantial speed up over the number of data frames collected and with the increase in sensors. 

In terms of future work, there are a lot of directions in which this framework could be extended. There is support given for defining novel sensors in CARLA, so using Car-STAGE to define a sensor such as flash LiDAR would be very useful for users. On the same vein, developing a module for a steering wheel command would prove to be beneficial to the AV research community. 

\addtolength{\textheight}{-12cm}   








\bibliographystyle{IEEEtran}
\bibliography{references}

\begin{thebibliography}{1}
\providecommand{\url}[1]{#1}
\csname url@samestyle\endcsname
\providecommand{\newblock}{\relax}
\providecommand{\bibinfo}[2]{#2}
\providecommand{\BIBentrySTDinterwordspacing}{\spaceskip=0pt\relax}
\providecommand{\BIBentryALTinterwordstretchfactor}{4}
\providecommand{\BIBentryALTinterwordspacing}{\spaceskip=\fontdimen2\font plus
\BIBentryALTinterwordstretchfactor\fontdimen3\font minus
  \fontdimen4\font\relax}
\providecommand{\BIBforeignlanguage}[2]{{%
\expandafter\ifx\csname l@#1\endcsname\relax
\typeout{** WARNING: IEEEtran.bst: No hyphenation pattern has been}%
\typeout{** loaded for the language `#1'. Using the pattern for}%
\typeout{** the default language instead.}%
\else
\language=\csname l@#1\endcsname
\fi
#2}}
\providecommand{\BIBdecl}{\relax}
\BIBdecl

\bibitem{dosovitskiy2017carla}
A.~Dosovitskiy, G.~Ros, F.~Codevilla, A.~Lopez, and V.~Koltun, ``Carla: An open
  urban driving simulator,'' in \emph{Conference on robot learning}.\hskip 1em
  plus 0.5em minus 0.4em\relax PMLR, 2017, pp. 1--16.

\bibitem{jang2021carfree}
J.~Jang, H.~Lee, and J.-C. Kim, ``Carfree: Hassle-free object detection dataset
  generation using carla autonomous driving simulator,'' \emph{Applied
  Sciences}, vol.~12, no.~1, p. 281, 2021.

\bibitem{nesti2022carla}
F.~Nesti, G.~Rossolini, G.~D'Amico, A.~Biondi, and G.~Buttazzo, ``Carla-gear: a
  dataset generator for a systematic evaluation of adversarial robustness of
  vision models,'' \emph{arXiv preprint arXiv:2206.04365}, 2022.

\bibitem{rosende2023urban}
S.~B. Rosende, D.~S.~J. Gavil{\'a}n, J.~Fern{\'a}ndez-Andr{\'e}s, and
  J.~S{\'a}nchez-Soriano, ``An urban traffic dataset composed of visible images
  and their semantic segmentation generated by the carla simulator,''
  \emph{Data}, vol.~9, no.~1, p.~4, 2023.

\bibitem{shah2018airsim}
S.~Shah, D.~Dey, C.~Lovett, and A.~Kapoor, ``Airsim: High-fidelity visual and
  physical simulation for autonomous vehicles,'' in \emph{Field and Service
  Robotics: Results of the 11th International Conference}.\hskip 1em plus 0.5em
  minus 0.4em\relax Springer, 2018, pp. 621--635.

\bibitem{mukhlas2020carla}
\BIBentryALTinterwordspacing
A.~Mukhlas, ``Carla 2d bounding box annotation module,'' in \emph{Online
  Resource}, 2020, accessed: 28-December-2023. [Online]. Available:
  \url{https://mukhlasadib.github.io/CARLA-2DBBox}
\BIBentrySTDinterwordspacing

\bibitem{astudillo2022mono}
A.~Astudillo, A.~Al-Kaff, and F.~Garc{\'\i}a, ``Mono-dcnet: Monocular 3d object
  detection via depth-based centroid refinement and pose estimation,'' in
  \emph{2022 IEEE Intelligent Vehicles Symposium (IV)}.\hskip 1em plus 0.5em
  minus 0.4em\relax IEEE, 2022, pp. 664--669.

\end{thebibliography}

\end{document}